# Important New Developments in Arabographic Optical Character Recognition (OCR)

By: Maxim Romanov, Matthew Thomas Miller, Sarah Bowen Savant, and Benjamin Kiessling

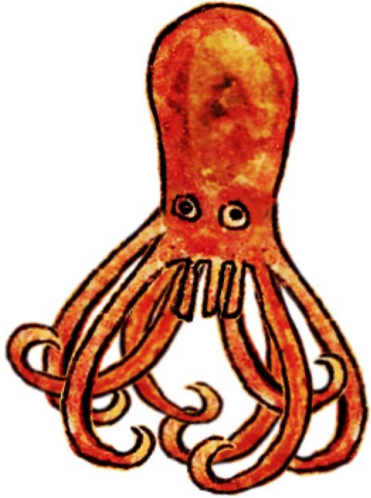

**Image**: *Kraken ibn Ocropus*, based on a depiction of an octopus from a manuscript of *Kitāb al-ʿashāʾish fī hāyūlā al-ʿilāj al-ṭibbī* (Leiden, UB : Or. 289); special thanks to Emily Selove for help with finding an octopus in the depths of the Islamic MS tradition.

## 1.1 Summary of Results of OpenITI's OCR

The **OpenITI** team[1]—building on the foundational open-source OCR work of the Leipzig University's (LU) Alexander von Humboldt Chair for Digital Humanities—has achieved Optical Character Recognition (OCR) accuracy rates for classical Arabic-script texts in the high nineties. These numbers are based on our tests of seven different Arabic-script texts of varying quality and typefaces, totaling over 7,000 lines (~400 pages, 87,000 words; see **Table 1** for full details). These accuracy rates not only represent a distinct improvement over the *actual*[2] accuracy rates of the various proprietary OCR options for classical Arabic-script texts, but, equally important, they are produced using an open-source OCR software called [Kraken](#) (developed by Benjamin Kiessling, LU), thus enabling us to make this Arabic-script OCR technology freely available to the broader Islamic, Persian, and Arabic Studies communities in the near future. In the process we also generated over

---

[1] The co-PIs of the Islamicate Texts Initiative (ITI) are Sarah Bowen Savant (Aga Khan University, London), Maxim G. Romanov (Leipzig University), and Matthew Thomas Miller (Roshan Institute for Persian Studies, University of Maryland, College Park).

[2] Proprietary OCR programs for Persian and Arabic (e.g., Sakhr's Automatic Reader, ABBYY Finereader, Readiris) over-promise the level of accuracy they deliver in practice when used on classical texts. These companies claim that they provide accuracy rates in the high 90 percentages (e.g., Sakhr claims 99.8% accuracy for high-quality documents). This may be the case for texts with simplified typeset and no short vowels; however, our tests of ABBYY Finereader and Readiris on high-quality scans of classical texts turned out accuracy rates of between 65% and 75%. Sakhr software was not available to us, as they offer no trial versions and it is the most expensive commercial OCR solution for Arabic. Moreover, since these programs are not open-source and offer only limited trainability (and created training data cannot be reused), their costs are prohibitive for most students and scholars and they cannot be modified according to the interests and needs of the academic community or the public at large. Most importantly, they have no web interfaces that would enable the production of wider, user-generated collections.

7,000 lines of "gold standard" (double-checked) data that can be used by others for Arabic-script OCR training and testing purposes.[3]

Table 1: Description of data

| Book* | Quality | Type | Size of data samples | | | |
|---|---|---|---|---|---|---|
| | | | Pages | Lines | Words | Chars |
| 0 Ibn al-Faqīh. *al-Buldān* | high*** | training | 79 | 1,466 | 16,909 | 92,730 |
| 1 Ibn al-Athīr. *al-Kāmil* | high*** | testing | 40 | 794 | 12,818 | 58,481 |
| 2 Ibn Qutayba. *Adab al-kātib* | high*** | testing | 55 | 794 | 7,848 | 42,230 |
| 3 al-Jāḥiẓ. *al-Ḥayawān* | high*** | testing | 65 | 992 | 11,870 | 59,191 |
| 4 al-Yaʿqūbī. *al-Taʾrīkh* | high*** | testing | 68 | 1,050 | 13,487 | 66,341 |
| 5 al-Dhahabī. *Taʾrīkh al-islām* | low**** | testing | 50 | 1,110 | 11,045 | 55,047 |
| 6 Ibn al-Jawzī. *al-Muntaẓam* | low**** | testing | 50 | 938 | 13,156 | 62,574 |
| | | **Total:** | 407 | 7,144 | 87,133 | 436,594 |

\* *Information on editions in at the end of the report*
\*\* *Performance on Arabic only (excluding punctuation and spaces)*
\*\*\* *300 dpi, grayscale; scanned specifically for the purpose of testing, with ideal parameters*
\*\*\*\* *200 dpi, black-and-white, pre-binarized; both downloaded from* www.archive.org *(via* www.waqfeya.org *)*

## 1.2 OCR and its Importance for Islamicate Studies Fields

Although there is a wealth of digital Persian and Arabic texts currently available in various open-access online repositories,[4] these collections still need to be expanded and supplemented in some important ways. OCR software is critical for this broader task of expanding the range of digital texts available to scholars for computational analysis. OCR programs, in the simplest terms, take an image of a text, such as a scan of a print book, and extract the text, converting the image of the text into a digital text that then can be edited, searched, computationally analyzed, etc.

The specific type of OCR software that we employed in our tests is an open-source OCR program called [Kraken](), which was developed by Benjamin Kiessling at Leipzig University's Alexander von Humboldt Chair for Digital Humanities. Unlike more traditional OCR approaches, [Kraken]() relies on a neural network—which mimics the way we learn—to recognize letters in the images of entire lines of text without trying first to segment lines into words and then words into letters. This segmentation step—a mainstream OCR approach that persistently fails on connected scripts—is thus completely removed from the process, making

---

[3] This gold standard data is available at: https://github.com/OpenArabic/OCR_GS_Data.
[4] Collecting and rendering these texts useful for computational textual analysis (through, for example, adding scholarly metadata and making them machine-actionable) is a somewhat separate but deeply interrelated project that OpenITI is currently working on as well.

[*Kraken*](#) uniquely powerful for dealing with the diverse variety of ligatures in connected Arabic script (see section 3.1 for more technical details).

**2.1 Initial OCR Tests**

We began our experiments by using [*Kraken*](#) to train a model[5] on high-quality[6] scans of ~1,000 lines of Ibn al-Faqīh's *al-Buldān* (work **#0**). We first generated training data (line transcriptions) for all of these lines, double checked them (creating so-called "gold standard" data), trained the model, and, finally, tested its ability to accurately recognize and extract the text. The results were impressive, reaching 97.56% accuracy for the entire text and an even more impressive 99.68% accuracy rate on the Arabic script alone (i.e., when errors related to punctuation and spaces were removed from consideration; such non-script errors are easy to fix in the post-correction phase and, in many cases, this correction process for non-script errors can be automated). See **Table 2**, *row #0* for full details.[7]

**Table 2: Accuracy Rates in Tests of our Custom Model**

|  |  |  | *Model accuracy level* | | | |
|---|---|---|---|---|---|---|
| **Book*** | **Quality** | **Type** | **Size 100** | **Ar\*\*** | **Size 200** | **Ar\*\*** |
| **0** Ibn al-Faqīh. *al-Buldān* | *high\*\*\** | *training* | 95.88 | 99.68 | 97.56 | 99.68 |
| **1** Ibn al-Athīr. *al-Kāmil* | *high\*\*\** | *testing* | 85.78 | 90.90 | 87.18 | 90.56 |
| **2** Ibn Qutayba. *Adab al-kātib* | *high\*\*\** | *testing* | 75.28 | 87.67 | 74.03 | 87.90 |
| **3** al-Jāḥiẓ. *al-Ḥayawān* | *high\*\*\** | *testing* | 69.03 | 72.78 | 68.32 | 71.87 |
| **4** al-Yaʿqūbī. *al-Taʾrīkh* | *high\*\*\** | *testing* | 78.78 | 83.42 | 78.28 | 81.85 |
| **5** al-Dhahabī. *Taʾrīkh al-islām* | *low\*\*\*\** | *testing* | 92.19 | 97.54 | 94.42 | 97.61 |
| **6** Ibn al-Jawzī. *al-Muntaẓam* | *low\*\*\*\** | *testing* | 90.40 | 97.39 | 92.26 | 97.80 |

    **\*** *Information on editions in at the end of the report*
  **\*\*** *Performance on Arabic only (excluding punctuation and spaces)*
 **\*\*\*** *300 dpi, grayscale; scanned specifically for the purpose of testing, with ideal parameters*
**\*\*\*\*** *200 dpi, black-and-white, pre-binarized; both downloaded from www.archive.org (via www.waqfeya.org)*

These numbers were so impressive that we decided to expand our study and use the model built on the text of Ibn al-Faqīh's *al-Buldān* (work **#0**) to OCR six other texts.

---

[5] "Training a model" is the term used in OCR work for teaching the OCR software to recognize a particular script or typeface—a process that only requires time and computing power. In our case, this process required 1 computer core and approximately 24 hours.
[6] "High quality" here means 300 dpi, grayscale images.
[7] We have also experimented with the internal configuration of our models: more extensive models, Size 200, showed slightly better accuracy in most cases (works #3-4 were an exception to this pattern), but it took twice as long to train and the OCR process using the larger model also takes more time.

We deliberately selected texts that were different from Ibn al-Faqīh's original text in terms of both their Arabic typeface/orthographic conventions and image quality. These texts represent at least two different typefaces (within which there are noticeable variations of font, spacing, and ligature styles), and four of the texts were high-quality scans while the other two were low-quality scans downloaded from www.archive.org (via www.waqfeya.org).[8]

When looking at the results in **Table 2**, it is important that the reader notes that works **#1-6** are "testing" data. That is, these accuracy results were achieved by utilizing a model built on the text of work **#0** to perform OCR on these other texts. For this reason it is not surprising that the accuracy rates for works **#1-4** are not as high as the accuracy rates for the training text, work **#0**. The point that is surprising is that the use of the work **#0**-based model on the low quality scans of works **#5-6** achieved a substantially higher accuracy rate (97.61% and 97.8% respectively on their Arabic script alone) than on the high-quality scans of works **#1-4**. While these higher accuracy rates for works **#5-6** are the result of a closer affinity between their typefaces and that of work **#0**, it also indicates that the distinction between high- and low-quality images is not as important for achieving high accuracy rates with *Kraken* as we initially believed. In the future, this will help reduce substantially both the total length of time it takes to OCR a work and the barriers to entry for researchers wanting to OCR the low-quality scans they already possess.

**Table 3: Ligature Variations in Typefaces**
*(the table highlights only a few striking differences and is not meant to be comprehensive; examples similar to those of the main text are "greyed out")*

| *Book* | | | | | | | | |
|---|---|---|---|---|---|---|---|---|
| **[#0]** Ibn al-Faqīh (d. 365/975). *al-Buldān* | لهم | بها | الملا | نج | لم | إلىٰ | فيها | نم |
| **[#1]** Ibn al-Athīr (d. 630/1232). *al-Kāmil fī al-taʾrīkh* | لهم | بها | الملا | نج | لم | إلى | فيها | نم |
| **[#2]** Ibn Qutayba (d. 276/889). *Adab al-kātib* | لهم | بها | Not Present in Text | نج | لم | إلى | فيها | نم |
| **[#3]** al-Jāḥiẓ (d. 255/868). *al-Ḥayawān* | لهم | بها | الملا | نج | لم | إلى | فيها | نم |
| **[#4]** al-Yaʿqūbī (d. 292/904). *al-Taʾrīkh* | لهم | بها | الملا | نج | لم | إلى | فيها | نم |
| **[#5]** al-Dhahabī (d. 748/1347). *Taʾrīkh al-islām* | لهم | بها | Not Present in Text | نج | لم | إلى | فيها | نم |
| **[#6]** Ibn al-Jawzī (d. 597/1201). *al-Muntaẓam* | لهم | بها | Not Present in Text | نج | لم | إلى | فيها | نم |

---

[8] "Low-quality" here means 200 dpi, black and white, pre-binarized images. In short, the standard quality of most scans available on the internet, which are the product of scanners that prioritize smaller size and speed of scanning for online sharing (i.e., in contrast to high-quality scans that are produced for long-term preservation).

The decreased accuracy results for works **#1-4** are explainable by a few factors:
(1) The typeface of works **#3-4** is different than work **#0** and it utilizes a number of ligatures that are not present in the typeface of work **#0** (for examples, see **Table 3** above).
(2) The typefaces of work **#1-2** are very similar to that of **#0**, but they both have features that interfere with the **#0**-based model. **#1** actually uses two different fonts, and the length of connections—*kashīda*s—between letters vary dramatically (between 0.3 *kashīda* to 2 *kashīda*s and everything in between), which is not the case with **#0**, where letter spacing is very consistent.
(3) The text of work **#2** is highly vocalized—it has more ḥarakāt than any other texts in the sample (and especially in comparison with the model work **#0**).
(4) The text of work **#2** also has very complex and overabundant punctuation with highly inconsistent spacing.

Our **#0**-based model could not completely handle these novel features in the texts of works **#1-4** because it was not trained to do so. As the results in **Table 4** of the following section show, new models can be trained to handle these issues successfully.

**Table 4: Accuracy Rates in Text-Specific Models**

| | | | *Model accuracy level* | |
|---|---|---|---|---|
| **Book*** | **Quality** | **Type** | **Size 100** | **Ar**** |
| **1** Ibn al-Athīr. *al-Kāmil* | *high**** | *training* | **93.79** | **97.71** |
| **2** Ibn Qutayba. *Adab al-kātib* | *high**** | *training* | **89.30** | **98.47** |
| **3** al-Jāḥiẓ. *al-Ḥayawān* | *high**** | *training* | **94.86** | **97.59** |
| **4** al-Yaʿqūbī. *al-Taʾrīkh* | *high**** | *training* | **96.81** | **99.18** |

   ***** *Information on editions in at the end of the report*
  ****** *Performance on Arabic only (excluding punctuation and spaces)*
 ******* *300 dpi, grayscale; scanned specifically for the purpose of testing, with ideal parameters*

**2.2 Round #2 Tests: Training New Models**

The most important advantage of [Kraken](#) is that its workflow allows one to train new models relatively easily, including text-specific ones. In a nutshell, the process of training requires a transcription of approximately 800 lines (the number will vary depending on the complexity of the typeface) aligned with images of these lines as they appear in the printed edition. The training itself takes 20-24 hours and is performed by a machine without human involvement; multiple models can be trained simultaneously. [Kraken](#) includes tools for the production of transcription forms (see **Figure 1** below); the data supplied through these forms is then used to train a new model. (Since there are a great number of Arabic-script texts that have already been converted into digital texts, one can use these as the base texts to fill in the forms more quickly—i.e., instead of typing the transcription—and then double-check them for accuracy; this was what we did, and it saved us a lot of time.)

**Figure 1: Kraken's Transcription Interface**

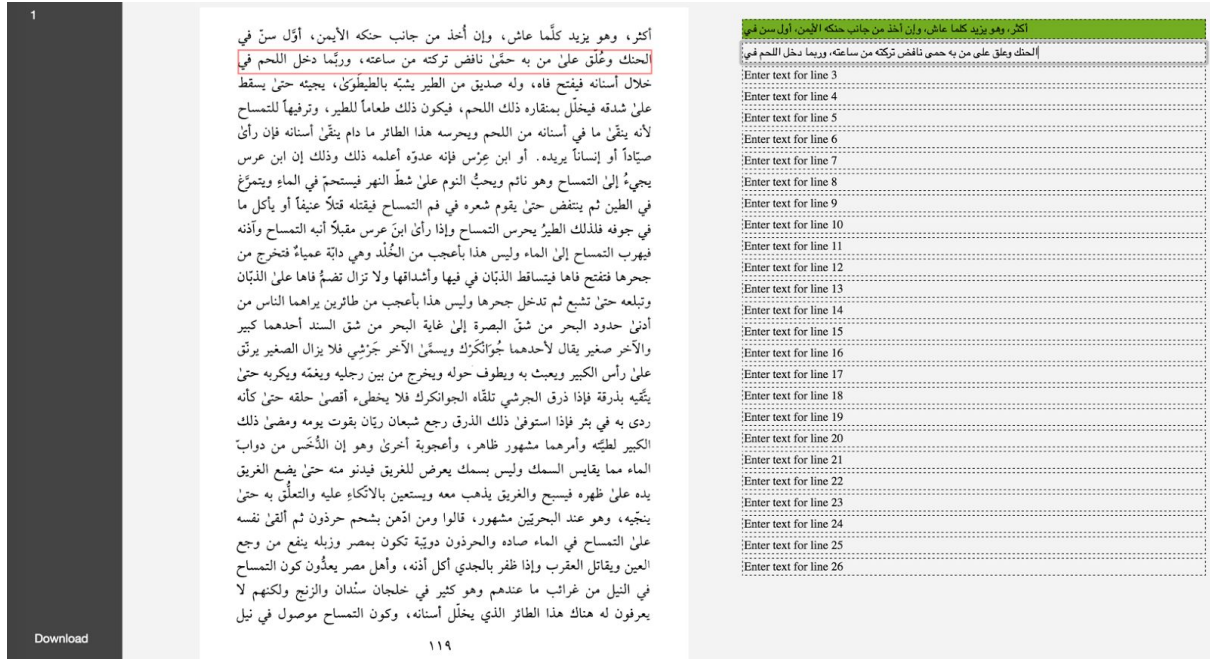

The importance of *Kraken's* ability to quickly train new models is illustrated clearly by texts such as works **#1-4** . When using the model built on work **#0** in our initial round of testing, we were only able to achieve accuracy rates ranging from the low seventies to low nineties on these texts (see **Table 2**). However, when we trained models on works **#1-4** specifically in our second round of testing, the accuracy rates for these texts substantially improved, reaching into the high nineties (see full results in **Table 4** above). The accuracy results for work **#4**, for example, improved from 83.42% on Arabic script alone in our first work #0-based model tests to 99.18% accuracy when we trained a mode on this text. The accuracy rates for works **#1-3** similarly improved, increasing from from 90.90% to 97.71% , 87.90% to 98.47%, from 72.78% to 97.59%, respectively. (See **Appendix** for the accuracy rates of these new models on all other texts as well.) These accuracy rates for Arabic-script recognition are already high, but we actually believe that they can improved further with larger training data sets.

Although the process of training a new model for a new text/typeface does require some effort, the only real time-consuming component is the generation of ~800 lines of gold standard line transcriptions. As we develop the OpenITI OCR project we will address the issue of the need for multiple models through a two-pronged strategy. First, we will try to train a general model, periodically adding new features that the model has not "seen" before. Secondly, we will train individual models for distinct typefaces and editorial styles (which sometimes vary in their use of vocalization, fonts, spacing, and punctuation), producing a library of OCR models that gradually will cover all major typefaces and editorial styles used in modern Arabic-script printing. There certainly are numerous Arabic-script typefaces and editorial styles that have been used throughout the last century and a half of Arabic-script printing, but ultimately the number is finite and definitely not so numerous as to make it impossible to create models for each over the long term.

**3.1 Conclusions and Next Steps for the OpenITI OCR Project**

The two rounds of testing presented here indicate that with a fairly modest amount of gold standard training data (~800–1,000 lines) [Kraken](.) is consistently able to produce OCR results for Arabic-script documents that achieve accuracy rates in the high nineties. In some cases, such as works **#5-6**, achieving OCR accuracy rates of up to 97.5% does not even require training a new model on that text. However, in other cases, such as works **#1-4**, achieving high levels of OCR accuracy does require training a model specific to that typeface, and, in some select cases of texts with similar typefaces but different styles of vocalization, font variations, and punctuation patterns (e.g., works **#1-2**), training a model for the peculiarities of a particular edition.

In the near future we are planning to develop a user-friendly web-interface for post-correction of the OCR output. Data supplied by users will allow us to train new models. It should be stressed that training edition-specific models is quite valuable, as there is a number of multivolume books—often with over a dozen volumes per text—that need to be converted into proper digital editions. In the long term, we will are also planning to train models for other Islamicate languages (Ottoman Turkish, Urdu, Syriac, etc.). Our hope is that an easy-to-use and effective OCR pipeline will allow us all—collectively—to significantly enrich our collection of digital Islamicate texts and thereby enable us to understand better this fascinating and understudied textual tradition.

**4.1 The Technical Details: *Kraken* and its OCR Method**

[Kraken](.) is the open-source OCR software that we used in our tests. Developed by Benjamin Kiessling at UL's Alexander von Humboldt Chair for Digital Humanities, [Kraken](.) is a "fork"[9] of the unmaintained *ocropus package*[10] combined with the CLSTM neural network library.[11] [Kraken](.) represents a substantial improvement over the *ocropus package*: its accuracy and performance rates are drastically better, it

---

[9] "Fork" is a computer-science term for a new independent development that builds on an existing software.
[10] For details, see: [https://github.com/tmbdev/ocropy](https://github.com/tmbdev/ocropy) and [https://en.wikipedia.org/wiki/OCRopus](https://en.wikipedia.org/wiki/OCRopus).
[11] See: [https://github.com/tmbdev/clstm](https://github.com/tmbdev/clstm).

supports right-to-left scripts and combined LTR/RTL (BiDi) texts, and it includes a rudimentary transcription interface for offline use.

The OCR method that powers [Kraken](#) is based on a long short-term memory (Hochreiter and Schmidhuber, 1997) recurrent neural network utilizing the Connectionist Temporal Classification objective function (Graves et al., 2006, as elaborated in Breuel et al., 2013). In contrast to other systems requiring character level segmentation before classification, it is uniquely suited for the recognition of connected Arabographic scripts because the objective function used during training is geared towards assigning labels—i.e., characters/glyphs—to regions of unsegmented input data.

The system works on unsegmented data both during training and recognition—its base unit is a text line (line recognizer). For training, a number of printed lines have to be transcribed using a simple HTML transcription interface (see **Figure 1** above). The total amount of training data, i.e. line image-text pairs, required may vary depending on the complexity of the typeface and number of glyphs used by the script. Acquisition of training data can be optimized by line-wise alignment of existing digital editions with printed lines, although even wholesale transcription is a faster and relatively unskilled task in comparison to training data creation for other systems such as *tesseract*.[12]

Our current models were trained on ~1,000 pairs each, corresponding to ~50-60 pages of printed text. Models are fairly typography specific, the most important factor being fonts and spacing, although some mismatch does not degrade recognition accuracy substantially (2-5%).[13] Thus new training data for an unknown typeface can be produced by correcting the output from a model for a similar font—in other words, generating training data for every subsequent model will require less and less time. Last but not least, it is also possible to train multi-typeface models by simply combining training data, albeit some parameter tuning is required to account for the richer typographic morphology that the neural network must learn.

## 5.1 Acknowledgements

We would never have been able to complete this work without the help of our team members at Leipzig University, University of Maryland (College Park), and Aga Khan University, London. We are deeply indebted to Benjamin Kiessling (Leipzig University) for his development of [Kraken](#) and for training models for Arabographic typefaces. We would also like to thank Elijah Cooke (Roshan Institute, UMD) for helping us to process the data, Samar Ali Ata (Roshan Institute, UMD) for--- generating several sets of high-quality scans for us, and Layal Mohammad (ISMC, AKU), Mohammad Meqdad (ISMC, AKU), and Fatemeh Shams (ISMC, AKU) for helping us to generate and double check the training data. Lastly, we would like to express our gratitude to Gregory Crane (Alexander von Humboldt Chair for Digital Humanities, LU), Fatemeh Keshavarz (Roshan Institute for Persian Studies, UMD), and David Taylor (ISMC, AKU) for their guidance and support of our work.

[12] See: https://github.com/tesseract-ocr and https://en.wikipedia.org/wiki/Tesseract_(software).

[13] For example, if a glyph is in a slightly different font than the one that the model was trained on, it may sometimes be misrecognized as another one (or not at all), thus leading the overall accuracy rate be slightly lower despite the fact that most of the other text is recognized correctly.

---

## Bibliography and Links

**Computer-Science Bibliography**
- Hochreiter, Sepp, and Jürgen Schmidhuber. "Long short-term memory." in *Neural computation* 9.8 (1997): 1735-1780.
- Graves, Alex, et al. "Connectionist temporal classification: labelling unsegmented sequence data with recurrent neural networks," in *Proceedings of the 23rd international conference on Machine learning*. ACM, 2006.
- Breuel, Thomas M., et al. "High-performance OCR for printed English and Fraktur using LSTM networks," in *12th International Conference on Document Analysis and Recognition*. IEEE, 2013.
- CLSTM neural network library: https://github.com/tmbdev/clstm

**Links to open source software:**
- *Nidaba:* https://openphilology.github.io/nidaba/
- *Kraken*: https://github.com/mittagessen/kraken
- *OCR-Evaluation tools*: https://github.com/ryanfb/ancientgreekocr-ocr-evaluation-tools

**OpenITI Gold-Standard Data for Arabic OCR:**
- Link: https://github.com/OpenArabic/OCR_GS_Data

**Editions of Printed Texts (when two dates are given, the second one is CE)**
- **[#0]** Ibn al-Faqīh (d. 365/975). *Al-Buldān*. Ed. Yūsuf al-Hādī. Beirut: ʿĀlam al-Kutub, 1996 CE.
- **[#1]** Ibn al-Athīr (d. 630/1232). *Al-Kāmil fī al-taʾrīkh*. Ed. ʿAbd Allāh al-Qāḍī. Beirut: Dār al-Kutub al-ʿIlmiyya, 1415/1994.
- **[#2]** Ibn Qutayba (d. 276/889). *Adab al-kātib*. Ed. Muḥammad al-Dālī. Muʾassasat al-Risāla, n.d.
- **[#3]** al-Jāḥiẓ (d. 255/868). *Al-Ḥayawān*. Beirut: Dār al-Kutub al-ʿIlmiyya, 1424/2003.
- **[#4]** al-Yaʿqūbī (d. 292/904). *Al-Taʾrīkh*. Beirut: Dār Ṣādir, n.d.

- **[#5]** al-Dhahabī (d. 748/1347). *Taʾrīkh al-Islām*. Al-Maktaba al-Tawfīqiyya, n.d.
- **[#6]** Ibn al-Jawzī (d. 597/1201). *Al-Muntaẓam*. Ed. Muḥammad al-Qādir ʿAṭā, Muṣṭafá al-Qādir ʿAṭā. Beirut: Dār al-Kutub al-ʿIlmiyya, 1412/1992.

---

## *Appendix*: Performance of New Models

### Table A: Performance of #1-Based Model on Other Texts

| Book* | Quality | Type | Model accuracy level | | | |
|---|---|---|---|---|---|---|
| | | | Size 100 | Ar** | Size 200 | Ar** |
| 1 Ibn al-Athīr. *al-Kāmil* | high*** | *training* | 93.79 | 97.71 | 93.58 | 97.59 |
| 2 Ibn Qutayba. *Adab al-kātib* | high*** | testing | 82.68 | 95.72 | 80.92 | 94.88 |
| 3 al-Jāḥiẓ. *al-Ḥayawān* | high*** | testing | 71.78 | 75.16 | 70.85 | 74.27 |
| 4 al-Yaʿqūbī. *al-Taʾrīkh* | high*** | testing | 79.67 | 84.40 | 78.12 | 82.21 |
| 5 al-Dhahabī. *Taʾrīkh al-islām* | low**** | testing | 90.68 | 95.95 | 90.37 | 95.78 |
| 6 Ibn al-Jawzī. *al-Muntaẓam* | low**** | testing | 93.33 | 98.51 | 92.96 | 98.22 |

\* *Information on editions in at the end of the report*
\*\* *Performance on Arabic only (excluding punctuation and spaces)*
\*\*\* *300 dpi, grayscale; scanned specifically for the purpose of testing, with ideal parameters*
\*\*\*\* *200 dpi, black-and-white, pre-binarized; both downloaded from www.archive.org (via www.waqfeya.org)*

### Table B: Performance of #2-Basel Model on Other Texts

| Book* | Quality | Type | Model accuracy level | | | |
|---|---|---|---|---|---|---|
| | | | Size 100 | Ar** | Size 200 | Ar** |
| 1 Ibn al-Athīr. *al-Kāmil* | high*** | testing | 83.52 | 88.56 | 83.55 | 88.56 |
| 2 Ibn Qutayba. *Adab al-kātib* | high*** | *training* | 89.30 | 98.47 | 89.42 | 98.44 |
| 3 al-Jāḥiẓ. *al-Ḥayawān* | high*** | testing | 74.82 | 76.51 | 74.87 | 76.65 |
| 4 al-Yaʿqūbī. *al-Taʾrīkh* | high*** | testing | 81.50 | 84.05 | 79.81 | 83.67 |
| 5 al-Dhahabī. *Taʾrīkh al-islām* | low**** | testing | 84.89 | 93.19 | 83.08 | 92.53 |
| 6 Ibn al-Jawzī. *al-Muntaẓam* | low**** | testing | 87.56 | 94.21 | 86.34 | 93.57 |

\* *Information on editions in at the end of the report*
\*\* *Performance on Arabic only (excluding punctuation and spaces)*
\*\*\* *300 dpi, grayscale; scanned specifically for the purpose of testing, with ideal parameters*
\*\*\*\* *200 dpi, black-and-white, pre-binarized; both downloaded from www.archive.org (via www.waqfeya.org)*

### Table C: Performance of #3-Basel Model on Other Texts

| Book* | Quality | Type | Model accuracy level | | | |
|---|---|---|---|---|---|---|
| | | | Size 100 | Ar** | Size 200 | Ar** |
| 1 Ibn al-Athīr. *al-Kāmil* | high*** | testing | 80.23 | 86.27 | 82.46 | 87.48 |
| 2 Ibn Qutayba. *Adab al-kātib* | high*** | testing | 80.90 | 91.54 | 82.61 | 93.24 |
| 3 al-Jāḥiẓ. *al-Ḥayawān* | high*** | *training* | 94.86 | 97.59 | 94.82 | 97.41 |
| 4 al-Yaʿqūbī. *al-Taʾrīkh* | high*** | testing | 90.91 | 95.13 | 91.28 | 94.71 |
| 5 al-Dhahabī. *Taʾrīkh al-islām* | low**** | testing | 81.93 | 91.23 | 83.03 | 92.22 |
| 6 Ibn al-Jawzī. *al-Muntaẓam* | low**** | testing | 84.07 | 93.58 | 86.26 | 94.20 |

 \* Information on editions in at the end of the report
 \*\* Performance on Arabic only (excluding punctuation and spaces)
 \*\*\* 300 dpi, grayscale; scanned specifically for the purpose of testing, with ideal parameters
 \*\*\*\* 200 dpi, black-and-white, pre-binarized; both downloaded from www.archive.org (via www.waqfeya.org)

### Table D: Performance of #4-Basel Model on Other Texts

| Book* | Quality | Type | Model accuracy level | | | |
|---|---|---|---|---|---|---|
| | | | Size 100 | Ar** | Size 200 | Ar** |
| 1 Ibn al-Athīr. *al-Kāmil* | high*** | testing | 79.80 | 86.35 | na | na |
| 2 Ibn Qutayba. *Adab al-kātib* | high*** | testing | 72.99 | 82.84 | na | na |
| 3 al-Jāḥiẓ. *al-Ḥayawān* | high*** | testing | 83.38 | 87.65 | na | na |
| 4 al-Yaʿqūbī. *al-Taʾrīkh* | high*** | *training* | 96.81 | 99.18 | na | na |
| 5 al-Dhahabī. *Taʾrīkh al-islām* | low**** | testing | 82.76 | 90.65 | na | na |
| 6 Ibn al-Jawzī. *al-Muntaẓam* | low**** | testing | 87.71 | 96.00 | na | na |

 \* Information on editions in at the end of the report
 \*\* Performance on Arabic only (excluding punctuation and spaces)
 \*\*\* 300 dpi, grayscale; scanned specifically for the purpose of testing, with ideal parameters
 \*\*\*\* 200 dpi, black-and-white, pre-binarized; both downloaded from www.archive.org (via www.waqfeya.org)